\documentclass[letterpaper, 10 pt, conference]{ieeeconf}  

\IEEEoverridecommandlockouts                              

\usepackage[utf8]{inputenc}

\usepackage[l2tabu,orthodox]{nag}

\usepackage[
    backend=bibtex8,
    style=ieee,
    sorting=none,
    natbib=true,
    doi=false,
    isbn=false,
    url=false,
    eprint=false,
    maxcitenames=1,
    mincitenames=1
]{biblatex}

\usepackage[pdftex,colorlinks]{hyperref}

\usepackage[printonlyused]{acronym}

\usepackage{siunitx}
\usepackage[all]{nowidow}

\usepackage[dvipsnames]{xcolor}

\usepackage{lipsum}

\usepackage{xspace} 
\newcommand{\ie}{i.e., }
\newcommand{\eg}{e.g., }

\usepackage[pdftex]{graphicx}

\usepackage{epstopdf}

\usepackage{import}

\graphicspath{{./latexGoodPractices/}}

\usepackage{booktabs}

\usepackage{tabularx}
\usepackage{multirow, multicol}

\usepackage{amssymb,amsfonts,amsmath,amscd}

\usepackage{bm}

\newcommand{\bbm}{\begin{bmatrix}}
\newcommand{\ebm}{\end{bmatrix}}

\usepackage{physics} 
\usepackage{amsmath}
\usepackage{amssymb}
\usepackage{stmaryrd}
\usepackage{mathtools}

\usepackage{standalone}
\usepackage{tikz}

\let\labelindent\relax
\usepackage{enumitem}

\usetikzlibrary{positioning}
\usetikzlibrary{shapes}

\newcommand{\nbruns}{32}
\newcommand{\maxangularspeed}{18.6}
\newcommand{\maxlinearacceleration}{157.8}

\newcommand{\maxangularspeedcollege}{3.6}
\newcommand{\maxlinearaccelerationcollege}{19.6}
\newcommand{\maxangularspeedhilti}{4.7}
\newcommand{\maxlinearaccelerationhilti}{30.7}
\newcommand{\maxlinearaccelerationimprovement}{127.1}
\newcommand{\maxangularspeedimprovement}{13.9}

\acrodef{DARPA}{Defense Advanced Research Projects Agency}
\acrodef{SLAM}{Simultaneous Localization And Mapping}
\acrodef{IMU}{Inertial Measurement Unit}
\acrodef{ICP}{Iterative Closest Point}
\acrodef{GICP}{Generalized Iterative Closest Point}
\acrodef{GF}{Gyro-free}
\acrodef{INS}{Inertial Navigation System}
\acrodef{EKF}{Extended Kalman Filter}
\acrodef{NDT}{Normal Distributions Transform}
\acrodef{IEKF}{Iterated Extended Kalman Filter}
\acrodef{COM}{Center Of Mass}
\acrodef{GP}{Gaussian Processe}
\acrodef{MEMS}{Micro-Electromechanical Systems}
\acrodef{LOAM}{Lidar Odometry And Mapping}
\acrodef{TW}{Time-based Weighting}
\acrodef{TIGS}{Tumbling-Induced Gyroscope Saturation}

\title{\LARGE \textbf{
		Saturation-Aware Angular Velocity Estimation: Extending the Robustness of SLAM to Aggressive Motions*
}}

\author{Simon-Pierre Deschênes$^{1}$, Dominic Baril$^{1}$, Mat\v ej Boxan$^{1}$, \\
	Johann Laconte$^{1}$, Philippe Giguère$^{1}$ and François Pomerleau$^{1}$
	\thanks{*This research was supported by the Fonds de recherche du Québec – Nature et technologies (FRQNT) and by the Natural Sciences and Engineering Research Council of Canada (NSERC) through grant CRDPJ 527642-18 SNOW (Self-driving Navigation Optimized for Winter). The authors wish to thank Benoît Audet for his help with the experimental setup.}
	\thanks{$^{1}$Northern Robotics Laboratory, Université Laval, Quebec City, Quebec, Canada
		{\texttt{\small \{simon-pierre.deschenes, francois.pomerleau\}@norlab.ulaval.ca}}}%
}

\newcommand\copyrighttext{%
	\scriptsize \copyright~2024 IEEE. Personal use of this material is permitted. Permission from IEEE must be obtained for all other uses, in any current or future media, including reprinting/republishing this material for advertising or promotional purposes, creating new collective works, for resale or redistribution to servers or lists, or reuse of any copyrighted component of this work in other works.}
\newcommand\copyrightnotice{%
	\begin{tikzpicture}[remember picture,overlay]
		\node[anchor=south,yshift=15pt] at (current page.south) {\parbox{\dimexpr\textwidth-\fboxsep-\fboxrule\relax}{\copyrighttext}};
	\end{tikzpicture}%
}

\makeatletter
\def\blx@err@patch#1{}
\makeatother

\begin{document}
	
	\maketitle
	\copyrightnotice
	\thispagestyle{empty}
	\pagestyle{empty}
	
	\begin{abstract}
		We propose a novel angular velocity estimation method to increase the robustness of \ac{SLAM} algorithms against gyroscope saturations induced by aggressive motions.
        Field robotics expose robots to various hazards, including steep terrains, landslides, and staircases, where substantial accelerations and angular velocities can occur if the robot loses stability and tumbles.
        These extreme motions can saturate sensor measurements, especially gyroscopes, which are the first sensors to become inoperative.
        While the structural integrity of the robot is at risk, the robustness of the \ac{SLAM} framework is oftentimes given little consideration.
        Consequently, even if the robot is physically capable of continuing the mission, its operation will be compromised due to a corrupted representation of the world.
        Regarding this problem, we propose a method to estimate the angular velocity using accelerometers during extreme rotations caused by tumbling.
        We show that our method reduces the median localization error by \SI[detect-weight,mode=text]{71.5}{\percent} in translation and \SI[detect-weight,mode=text]{65.5}{\percent} in rotation and is robust to mapping failures, which occurred in \SI[detect-weight,mode=text]{37.5}{\percent} of the experiments without our method.
        We also propose the \ac{TIGS} dataset, which consists of outdoor experiments recording the motion of a mechanical lidar subject to angular velocities four times higher than other similar datasets available.
        The dataset is available online at \url{https://github.com/norlab-ulaval/Norlab_wiki/wiki/TIGS-Dataset}.
	\end{abstract}
	
	\acresetall 
        \section{INTRODUCTION}~\label{sec:into}
\begin{figure}[t]
	\centering
	\includegraphics[width=0.48\textwidth]{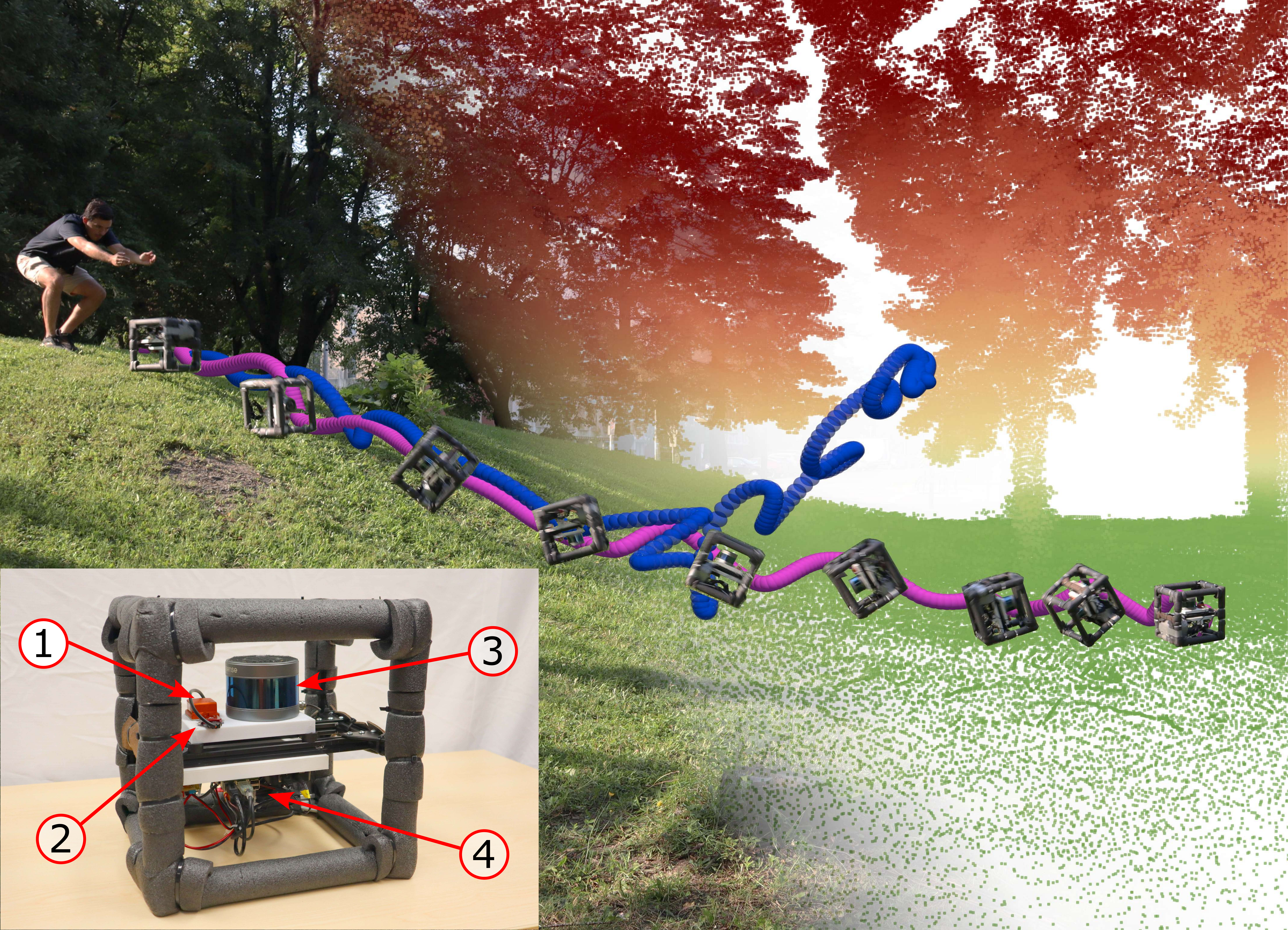} 
	\caption{Our robot localization system tumbling down a steep hill.
    At the top is a picture of the event and the reconstructed point cloud.
    The blue trajectory represents a~\ac{SLAM}-estimated trajectory relying on raw gyroscope measurements.
    In pink is a similar trajectory, which is estimated relying on our angular velocity estimation approach.
    The rugged perception rig is shown in the bottom left.
    The numbers in the red circles correspond to (1) XSens MTi-30~\ac{IMU}, 
    (2) VectorNav VN-100~\ac{IMU},
    (3) RoboSense RS-16 lidar,
    and (4) Raspberry~Pi~4.
 }
	\label{fig:motivation}
\end{figure}

	For many robot applications, operations are conducted in a remote, or dangerous environment, where human intervention is impossible~\citep{Nagatani2013}.
Hardware improvements have significantly reduced potential failure due to collisions, especially for aerial systems~\citep{Dilaveroglu2020}.
However, software systems, particularly robot localization, will typically not recover from falls, drops, and collisions~\citep{Ebadi2023}.
Therefore, increasing mobile robot localization robustness to such events is key to enabling autonomy in human-denied environments.
Inspired by work on control, such as \citet{Williams2018}, we define aggressive motions for perception as being near the dynamic limits that the system can sustain.
With this definition, navigation on highways would not cause aggressive motions despite high velocities.
On the contrary, a robot tumbling down a steep hill, as shown in the top part of \autoref{fig:motivation}, exemplifies this definition of aggressive motions very well because of the repeated collisions and fast angular velocities that are sustained.
Such motions cause skew in lidar scans~\cite{Deschenes2021} and saturation in gyroscope measurements~\cite{Lee2019}.
Deskewing algorithms correct these distortions using an estimate of the intra-scan lidar motion.
However, in many \ac{SLAM} systems, the prior attitude for optimization and estimate for intra-scan lidar motion is obtained by integrating \ac{IMU} measurements \cite{Shan2020, Reinke2022, Xu2022, Chen2023}.
Therefore, gyroscope saturations lead to wrong optimization priors, inaccurate deskewing, and thus, to \ac{SLAM} failure, as shown by the blue trajectory.
In this work, we propose to leverage the theory related to \ac{GF} \ac{INS}~\cite{Pachter2013} to estimate angular velocities during gyroscope saturations.
To validate our approach without damaging robots, we built a rugged lidar-inertial rig, shown in the bottom left.
We generated a dataset of the rig tumbling down a steep hill, saturating gyroscope measurements to analyze our solution.
Thus, the contributions of this work are:
\begin{enumerate}
    \item a novel method to estimate robot angular velocities during gyroscope saturation periods;
    \item the \ac{TIGS} dataset, consisting of \nbruns~distinct runs of a custom perception rig tumbling down a steep hill, reaching angular velocities up to~\SI{\maxangularspeed}{\radian/\second}.
\end{enumerate}

	\section{RELATED WORK}~\label{sec:rw}
	In this section, we describe recent work in the literature focused on localization and mapping under aggressive motions. 
In particular, we explain how these approaches were not tested or would not function in cases where gyroscope saturations occur during aggressive motions.
Then, we describe existing~\ac{GF}-\ac{INS} methods, aiming to estimate the angular velocity of a robot when gyroscope measurements are saturated.
Lastly, we analyze mechanical lidar \ac{SLAM} datasets and demonstrate that they are not suited to test our angular velocity estimation method.

\subsubsection{SLAM robust to aggressive motions}~\label{sec:rw-slam}
Several \ac{SLAM} algorithms were proposed to overcome the challenges posed by aggressive motions.
In the FAST-LIO2 \ac{SLAM} algorithm~\citep{Xu2022}, an~\ac{IEKF} back-propagates the estimated state to deskew the point cloud after the prediction step.
FAST-LIO2 was tested at angular speeds up to \SI{21.7}{\radian/\second}, without specifying accelerations and with no mention of gyroscope saturations.
Another algorithm robust to aggressive motions is the DLIO \ac{SLAM} algorithm \cite{Chen2023}.
In DLIO, scans are deskewed using the lidar motion estimated by integrating \ac{IMU} measurements with a constant jerk and angular acceleration model.
After roughly aligning the scan with the map through deskewing, the scan alignment is refined using the \ac{GICP} \cite{Segal2009} registration algorithm.
Their method was tested at angular velocities up to \SI{\maxangularspeedcollege}{\radian/\second} and linear accelerations up to \SI{\maxlinearaccelerationcollege}{\meter/\square\second}, but was not tested under saturated gyroscope measurements.
Although promising, the aforementioned methods use \ac{IMU} measurements to compute the prior for their optimization process.
If \ac{IMU} measurements are incomplete because of saturations, they might lead the optimization to converge far from the true solution.
An approach to tackle sensor failures is introduced in the LOCUS \ac{SLAM} algorithm \cite{Palieri2021}.
They introduce a health-monitoring module in their method to detect sensor malfunctions.
In contrast, we propose an approach that not only detects but also recovers from gyroscope failures, as robots do not have a direct alternative for such measurements.
To our knowledge, the only lidar-inertial \ac{SLAM} framework which is robust to \ac{IMU} saturations is Point-LIO \cite{He2023}.
It consists of an on-manifold \ac{EKF} that registers each individual point to the closest plane as it is measured and that uses a kinematic model to model \ac{IMU} measurements as an output.
Point-LIO therefore does not need to deskew incoming scans and is more robust to \ac{IMU} saturations.
To test their method, the authors conducted experiments in which gyroscope saturations were encountered over smooth motions.
In our previous work \cite{Deschenes2021}, we introduced a \ac{SLAM} algorithm that takes into account the skewing uncertainty during registration.
This allowed our localization and mapping algorithm to give more importance to certain portions of a scan that were less affected by scan skewing.
However, since our method was not robust to gyroscope saturation, we limited our experiments to angular speeds up to \SI{11}{rad/\second} and linear accelerations up to \SI{200}{\meter/\second}$^2$.
The aforementioned limitations motivate the need for an angular velocity estimation method relying on other sensory measurements, which can be relied upon in the case of gyroscope saturations during aggressive motions.

\subsubsection{Angular Velocity Estimation}~\label{sec:rw-speed-estimation}
Several solutions have been proposed to estimate gyroscope measurements during saturation periods.
In the work of \citet{Dang2014}, the authors propose a smoothing algorithm to estimate saturated gyroscope measurements.
They use an optimization algorithm based on the presence of zero-velocity intervals for motion tracking.
Their method is well-suited in situations in which short gyroscope-saturated time windows occur during a continuous motion contained between zero-velocity periods.
However, their method was not designed for cases where repeated collisions are sustained (\eg when tumbling).
Alternatively, \citet{Tan2020} introduce an \ac{EKF} exploiting the sinusoidal structure of magnetometer measurements to estimate the angular velocity of a monocopter, despite gyroscope saturations.
In a situation where repeated collisions are sustained, a sinusoidal structure in magnetometer measurements cannot be assumed.
Moreover, in robotics, magnetometers are often disregarded as their measurements are biased by proximal magnetic sources \cite{Silic2020}.
Another approach is explored in the work of \citet{Pachter2013}, where \ac{GF} \ac{INS} theory is applied to allow the estimation of the position, orientation, linear velocity, and angular velocity of an object in 3D using only accelerometers.
Following this work, \citet{Lee2019} proposed an \ac{EKF} to estimate the angular velocity of a rotating plate using three accelerometers, which they validated experimentally.
This solution was developed for aerospace applications and was not tested inside a \ac{SLAM} framework.
Since accelerometer-based methods have more potential than other work presented previously, we will build on these solutions to improve the robustness of~\ac{SLAM} algorithms under saturated gyroscope measurements.

\subsubsection{Aggressive Motion Datasets}
In order to demonstrate the improvement of \ac{SLAM} reached through our speed estimation method, a dataset with aggressive motions and gyroscope saturations is required.
We studied the mechanical lidar \ac{SLAM} datasets that are most commonly used and that contain the most aggressive motions, namely the Newer College \cite{Ramezani2020} and Hilti-Oxford \cite{Zhang2023} datasets.
Because of its importance in the literature, we also studied the KITTI dataset \cite{Geiger2012}.
The maximum angular velocity in all of these datasets combined is \SI{\maxangularspeedhilti}{\radian/\second} and the maximum linear acceleration in all datasets combined is \SI{\maxlinearaccelerationhilti}{\meter/\square\second}.
Since the motions in these datasets are not aggressive enough to cause gyroscope saturations, we propose the \acf{TIGS} dataset, which consists of a perception rig tumbling down a hill, with angular velocities up to \SI{\maxangularspeed}{\radian/\second} and linear accelerations up to \SI{\maxlinearacceleration}{\meter/\square\second}.

	\section{THEORY}~\label{sec:theory}
	To increase the robustness of \ac{SLAM} algorithms to a robot tumbling or colliding with its environment, we develop a method that allows the estimation of the angular velocity of a robot when its gyroscope measurements are saturated.
The only prerequisite of our angular velocity estimation method is an estimate of the robot's \ac{COM} location.
We provide the uncertainty of the estimated velocity to allow its use in probabilistic frameworks (\eg Bayesian filtering).
We then describe the \ac{SLAM} framework in which our method is inserted.
Our angular velocity estimation method implementation is freely available online to facilitate replicability.\footnote{\url{https://github.com/norlab-ulaval/saturated_gryo_speed_estimation}}

\subsection{Angular Velocity Estimation}~\label{sec:speed-estimation}
\begin{figure}[htbp]
	\centering
	\includegraphics[width=\linewidth]{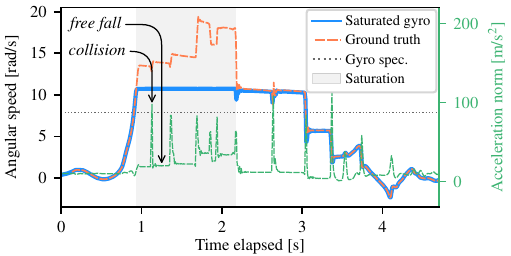}
        \caption{The angular speed is shown through time for the saturated gyroscope axis of a robot tumbling down a hill.
        Light gray zones indicate the gyroscope saturation periods.
        The measurements from a saturated gyroscope are shown in blue, the ground-truth angular speeds are indicated in orange, and the norm of the measured acceleration is indicated in dashed green.
        We show the manufacturer-specified gyroscope saturation point in dark gray.
        Examples of collisions with the ground and free-fall events are highlighted.}
	\label{fig:angular-velocity-curve}
\end{figure}

Inertial measurements during an event of a robot tumbling down a hill are shown in~\autoref{fig:angular-velocity-curve}.
Gyroscope saturations usually occur during the middle section of the tumbling, when the angular velocities are at their highest.
Accelerometers, on the other hand, tend to saturate less and, even if they do saturate, it is for a short period (\eg during a collision), as opposed to gyroscopes, which can saturate for several seconds.
We therefore have two distinct cases during which to estimate saturated gyroscope measurements: \emph{i)} during free-fall and \emph{ii)} during collisions.
Indeed, the plateaus in the angular speed curve correspond to free-fall periods whereas the fast changes correspond to collisions, as indicated by the spikes in the acceleration curve.

Since the modeling of collisions is still an open problem, it is challenging to estimate the angular velocity from accelerometer measurements during collisions.
Therefore, our approach is split into two steps.
First, we estimate the angular velocities assuming free-fall conditions.
Then, to account for the broken free-fall assumptions during collisions, we smooth the estimated velocities with a physically-motivated motion prior.
As shown in~\autoref{fig:angular-velocity-curve}, collisions are much shorter than free-fall periods, indicating that the robot is indeed in free fall during most of the gyroscope saturation period.
The following assumptions are made to estimate saturated gyroscope measurements during free fall:
\begin{enumerate}[label=\textbf{Assumption~\arabic*:}, ref=\arabic*, align=left, leftmargin=20pt, rightmargin=20pt, itemindent=0pt]
    \item The \ac{IMU} is not located along the robot's rotation axis;~\label{hyp:rotation-axis-imu}
	\item The measured linear acceleration at the robot's \ac{COM} is null;~\label{hyp:null-acceleration}
    \item The rotation axis remains unchanged between two \ac{IMU} measurements;~\label{hyp:unchanging-rotation-axis}
	\item The rotation axis passes through the robot's \ac{COM};~\label{hyp:rotation-axis-com}
	\item Only one axis of the gyroscope is saturated at once.~\label{hyp:single-axis-saturation}
\end{enumerate}
\autoref{hyp:rotation-axis-imu} is necessary to enable angular velocity estimation from the measured centripetal acceleration.
\autoref{hyp:null-acceleration} stems from the fact that the acceleration perceived by a body in free fall is null and on the underlying assumption that the force caused by air friction is negligible.
\autoref{hyp:unchanging-rotation-axis} is supported by the high acquisition rate of \ac{IMU} measurements, which is typically \SI{100}{Hz} or more, and by the angular momentum preventing the axis of rotation from changing quickly.
\autoref{hyp:rotation-axis-com} relies on the fact that when no external forces act on a body, it rotates about its \ac{COM}. 
This is again based on the underlying assumptions of free-fall conditions and of negligible air friction.
\autoref{hyp:single-axis-saturation} is not strictly necessary, but allows us to compute a simple and precise estimate of the angular speed of a saturated gyroscope axis.
Moreover, in our experiments described in \autoref{sec:setup}, we did not encounter situations in which more than one gyroscope axis was saturated at once.

\begin{figure}[htbp]
    \vspace{0.15cm}
	\centering
	\includegraphics[width=.7\linewidth]{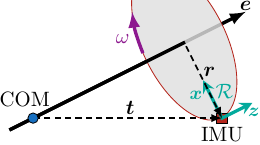} 
	\caption{Illustration of important quantities in our angular velocity estimation method. The \ac{COM} is indicated by a blue dot. The \ac{IMU} is indicated by a red square. The axis of rotation $\bm{e}$ is assumed to pass through the robot's \ac{COM}. The vector $\bm{t}$ joins the \ac{COM} and the \ac{IMU} and the rotation lever arm $\bm{r}$ joins the axis of rotation to the \ac{IMU}. The angular speed $\omega$ of the \ac{IMU} around $\bm{e}$ is indicated in purple. The $x$ and $z$ axis of the rotational frame $\mathcal{R}$ are illustrated in green.}
	\label{fig:speed-estimation}
\end{figure}

The important variables are illustrated in \autoref{fig:speed-estimation} where an \ac{IMU} is linked to the robot's \ac{COM} by $\bm{t}$ and rotates at an angular speed $\omega = \lVert \bm{\omega} \rVert$ around the unit rotation axis $\bm{e}$.
The $\bm{r}$ vector orthogonally links $\bm{e}$ to the \ac{IMU}.
The rotational coordinate frame $\mathcal{R}$ is at the same location as the \ac{IMU}, but rotated to have its $x$ axis perpendicular and pointing to the rotation axis $\bm{e}$ and its $z$ axis in the same direction as $\bm{e}$.
As the axis-angle representation states, the rotation axis $\bm{e}$ can be recovered from the angular velocity $\bm{\omega} = [\omega_x, \omega_y, \omega_z]^T$ such that $\bm{\omega} = \omega \bm{e}$.
Drawing from the work of \citet{Pachter2013}, the Coriolis formula states that
\begin{equation}
	\bm{a}_I = \bm{a}_C + \dot{\bm{\omega}} \cross \bm{r} + \bm{\omega} \cross (\bm{\omega} \cross \bm{r}),
\end{equation}
where $\bm{a}_I$ is the linear acceleration at the location of the \ac{IMU}, $\bm{a}_C$ is the linear acceleration at the robot's \ac{COM}, and $\dot{\bm{\omega}}$ is the angular acceleration of the robot.
All angular velocity and linear acceleration measurements are expressed in a common coordinate frame.
Since accelerometers measure proper acceleration, the measured acceleration $\tilde{\bm{a}}_I$ at the location of the \ac{IMU} is equal to
\begin{equation}~\label{eq:complicated-acc-measurement}
	\begin{aligned}
		\tilde{\bm{a}}_I &= \bm{a}_I - \bm{g} \\
		&= (\bm{a}_C - \bm{g}) + \dot{\bm{\omega}} \cross \bm{r} + \bm{\omega} \cross (\bm{\omega} \cross \bm{r}) \\
		&= \tilde{\bm{a}}_C + \dot{\bm{\omega}} \cross \bm{r} + \bm{\omega} \cross (\bm{\omega} \cross \bm{r}),
	\end{aligned}
\end{equation}
where $\bm{g}$ is the gravity force vector and $\tilde{\bm{a}}_C$ is the measured acceleration at the robot's \ac{COM}.
Using \autoref{hyp:null-acceleration}, \autoref{eq:complicated-acc-measurement} simplifies to
\begin{equation}~\label{eq:simplified-acc-measurement}
	\tilde{\bm{a}}_I \approx \dot{\bm{\omega}} \cross \bm{r} + \bm{\omega} \cross (\bm{\omega} \cross \bm{r}).
\end{equation}
\autoref{eq:simplified-acc-measurement} is the key to allowing the computation of angular velocity during gyroscope saturation periods.
The first term of the sum is the tangential acceleration of the \ac{IMU} and is oriented into the page in \autoref{fig:speed-estimation}.
The second term of the sum is the centripetal acceleration and is oriented in the same direction as the $x$ axis of the rotational coordinate frame $\mathcal{R}$ in \autoref{fig:speed-estimation}.
Therefore, expressing the accelerometer measurements $\tilde{\bm{a}}_I$ in the coordinate frame $\mathcal{R}$ and using \autoref{eq:simplified-acc-measurement}, we can deduce without further approximation that
\begin{equation}~\label{eq:vector-to-scalar}
	{}^{\mathcal{R}}\tilde{\bm{a}}_I
    \approx 
    \begin{bmatrix}
        \lVert \bm{\omega} \cross (\bm{\omega} \cross \bm{r}) \rVert \\
        -\lVert \dot{\bm{\omega}} \cross \bm{r} \rVert \\
        0
    \end{bmatrix} = \begin{bmatrix}
        \omega^2 r \\
        -\dot{\omega} r \\
        0
    \end{bmatrix},
\end{equation}
where $\omega = \lVert \bm{\omega} \rVert$, $r = \lVert \bm{r} \rVert$ and $\dot{\omega} = \lVert \dot{\bm{\omega}} \rVert$.
The last equality in \autoref{eq:vector-to-scalar} holds because $\bm{r}$ is orthogonal to $\bm{\omega}$ by definition and to $\dot{\bm{\omega}}$ due to \autoref{hyp:unchanging-rotation-axis}.
The $y$ component of ${}^{\mathcal{R}}\tilde{\bm{a}}_I$ is negative because the $y$ axis of $\mathcal{R}$ and the \ac{IMU} tangential acceleration are in opposite directions.
From here, the angular velocity can be estimated from either the $x$ or $y$ component of the acceleration vector.
However, computing the angular velocity via the angular acceleration $\dot{\bm{\omega}}$ would lead to integrating noise and thus lead to a less accurate estimate.
In order to compute the magnitude of the angular velocity vector $\lVert \bm{\omega} \rVert$, the magnitude of the lever arm $\lVert \bm{r} \rVert$ must be determined.
Using \autoref{hyp:rotation-axis-com}, as can be seen in \autoref{fig:speed-estimation}, $\bm{r}$ can be retrieved with
\begin{equation}
	\bm{r} = \bm{t} - (\bm{t} \cdot \bm{e}) \bm{e}.
\end{equation}
The axis of rotation $\bm{e}$ is usually determined using the angular velocity $\bm{\omega}$, but this is not possible in the present case, since the measurement of one of the gyroscope axes is saturated.
Using \autoref{hyp:unchanging-rotation-axis}, the axis of rotation of the previous estimated angular velocity is used instead.
Lastly, without loss of generality, let us assume that the gyroscope is saturated on the $x$ component.
Using \autoref{hyp:rotation-axis-imu} and \autoref{hyp:single-axis-saturation}, we can retrieve the saturated measurement $\omega_x$ using
\begin{equation}~\label{eq:final-equation}
    \omega_x = \sqrt{\frac{\tilde{a}_x}{\lVert \bm{t} - (\bm{t} \cdot \bm{e}) \bm{e} \rVert} - \omega_y^2 - \omega_z^2},
\end{equation}
where $\tilde{a}_x$ is the $x$ component of ${}^{\mathcal{R}}\tilde{\bm{a}}_I$, $\omega_y$ and $\omega_z$ are the unsaturated gyroscope measurements.
Due to the noise in accelerometer measurements, the computed angular speed might be below the saturation point, which is not possible.
To solve this, we conserve the maximum between the estimated angular speed magnitude and the saturation point.
The sign ambiguity of the computed angular speed can be resolved by considering the sign of the saturated gyroscope measurement.
Again, due to the noise in accelerometer measurements, the term under the radical in \autoref{eq:final-equation} can be negative.
In that case, we simply reject the estimate.
We are left with \autoref{eq:final-equation} to estimate the angular velocity when a saturation is detected using a threshold on gyroscope measurements.

We now smooth the angular velocity estimates computed previously with \acp{GP} using a physically-motivated motion prior.
\acp{GP} allow us to obtain more accurate angular velocity estimates during collisions, when our free-fall assumptions are broken.
Similarly to what was done by~\citet{Tang2019}, a white-noise-on-jerk motion prior is used.
To account for the possibly abrupt changes in angular velocity, the diagonal entries of the angular jerk power spectral density matrix are set to a high value $q_{\ddot{\omega}}$.
The unsaturated gyroscope measurements are assigned a covariance of $\sigma_{\tilde{\omega}}^2$, which is computed using the \ac{IMU} specifications.
The valid angular speed estimates are given a higher covariance, $\sigma_{\hat{\omega}}^2$, which is a parameter of our method.
Employing \acp{GP} for smoothing has the advantage of yielding both the mean and covariance of the estimated angular velocity as functions of time.
The STEAM library, from \citet{Anderson2015}, was used to carry out these computations.

\subsection{SLAM framework}~\label{sec:slam-theory}
The \ac{SLAM} framework in which our angular velocity estimation method is inserted is divided into four steps which are described briefly in this section.
For more details, refer to our previous work~\citep{Deschenes2021}.
\textbf{1)~Intra-scan trajectory estimation:} Using the estimated angular velocities and accelerometer measurements, the trajectory of the \ac{IMU} is estimated.
To do so, first, the angular velocities and linear accelerations are passed through a Madgwick filter \cite{Madgwick2011} to estimate the orientation of the \ac{IMU} throughout the scan.
These orientations are used to remove the gravity vector from accelerometer measurements.
Then, accelerometer measurements are integrated, and resulting displacements are added to the position computed in the previous registration to estimate the position of the \ac{IMU} throughout the scan.
Finally, this position and orientation information is used in combination with the extrinsic calibration between the~\ac{IMU} and lidar to compute the trajectory of the latter during the scan.
\textbf{2)~Deskewing:} As lidar sensors typically assume they are static during scans~\citep{Deschenes2021}, point positions need to be corrected with respect to intra-scan motion.
The poses of the lidar previously estimated at each \ac{IMU} measurement time are linearly interpolated to transform every measured point in the coordinate frame of the lidar at the beginning of the scan.
\textbf{3)~Uncertainty-aware registration:} Using the \ac{TW} model and registration algorithm described in \cite{Deschenes2021}, the deskewed scans are registered to the reconstructed map of the environment.
This weighting model takes the uncertainty of the deskewing into account for the registration algorithm by assigning a larger weight to scan points that are likely to be less affected by skewing.
As the estimated displacement of the lidar during the scan is used as a prior alignment for the registration algorithm, the quality of \ac{IMU} measurements has a major influence on the robustness of registration.
\textbf{4)~Merge and map maintenance:} The registered deskewed scan is then merged into the map of the environment and maintenance operations are performed.
These maintenance operations are surface normal computation and removal of points with a deskewing uncertainty above $\sigma_p^2$.

	
	\section{RESULTS}~\label{sec:results}
	In this section, we describe the experimental setup used to build our dataset. 
We show that our angular speed estimation method significantly reduces the angular velocity error in the case of gyroscope-saturating motions.
We then show the robustness improvement for our~\ac{SLAM} framework, both for localization and mapping.
Lastly, we compare the range of motions in our dataset to those in other \ac{SLAM} datasets.

\subsection{Experimental setup}~\label{sec:setup}
To validate the improvements reached through our approach while minimizing hardware damage and replacement costs, we created a rugged perception rig, which is shown in~\autoref{fig:motivation}.
A RoboSense RS-16 lidar was used to record the 3D point clouds at a frequency of~\SI{10}{\hertz}.
For angular velocity measurements, we used two different~\acp{IMU}, with distinct gyroscope saturation points.
The first \ac{IMU} is an XSens MTi-30, with a gyroscope saturating at~\SI{10.5}{\radian/\second}, despite the Xsens specification sheet stating a saturation point of \SI{7.85}{\radian/\second}.
The second \ac{IMU} is a VectorNav VN-100, with a gyroscope saturating at~\SI{34.9}{\radian/\second} according to its specification sheet.
Its angular velocity measurements are used as ground truth since we did not reach its gyroscope saturation point in our dataset.
Lastly, we used a Raspberry Pi 4 embedded computer to record all sensor data.
All~\ac{SLAM} results were computed offline, using the~\texttt{norlab\_icp\_mapper} library~\citep{Baril2022}.
The~\ac{COM} of the rig was evaluated manually, by balancing the rig on a single point on each face.
The constants that were introduced in \autoref{sec:theory} are set to $q_{\ddot{\omega}} = 10^6$, $\sigma_{\tilde{\omega}}^2 = 2.74\times10^{-5}$, $\sigma_{\hat{\omega}}^2 = 3.65$ and $\sigma_p^2 = 1$.
The value of $\sigma_{\tilde{\omega}}^2$ was computed from the MTi-30 datasheet and the values of $q_{\ddot{\omega}}$ and $\sigma_p^2$ are hyperparameters of our method.
To evaluate our method, we built the \ac{TIGS} dataset, including a total of \nbruns~distinct runs, consisting of pushing the rig to roll down a steep hill, mimicking a tumbling robot.
One of the runs of our dataset can be seen in~\autoref{fig:motivation}.
A ground-truth map was built by moving the sensor rig slowly, thus limiting skew in the scans.

\subsection{Angular velocity estimation}~\label{sec:result-angular-velocity}
\begin{figure}[htbp]
	\centering
	\includegraphics[width=\linewidth]{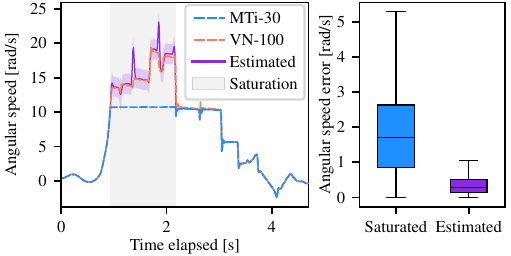}
        \caption{The left plot shows an example of the angular speed through time for the saturated gyroscope axis for a single run of our experiments.
        The measurements from an MTi-30 gyroscope are shown in dashed blue, the measurements from a VN-100 gyroscope are indicated in dashed orange, and the angular speeds estimated with our method using MTi-30 measurements are illustrated in purple.
        The purple-shaded area represents three standard deviations above and below the estimated speed. 
        The right plot shows the error in angular speed without (in blue) and with (in purple) our method during saturation periods for all runs.}
   	\label{fig:angular-velocity-error}
\end{figure}

Using the angular velocity estimation method described in \autoref{sec:speed-estimation}, the angular velocity of the platform was estimated for all of the runs in our dataset.
An example is shown for a single run in the left subplot of \autoref{fig:angular-velocity-error}.
Our method was applied to the MTi-30 measurements and the VN-100 measurements were used as ground truth.
The speed estimation error during gyroscope saturation periods with and without our method is shown for all runs in our dataset in the right part of \autoref{fig:angular-velocity-error}.
Only periods of saturation are studied (\ie the light gray area), as angular velocities are the same with or without our speed estimation method outside the saturation zones.
When accounting for all runs, our approach reduces the angular velocity error median by~\SI{83.4}{\%}, when compared to saturated gyroscope measurements.
As expected, our angular velocity estimation approach significantly reduces the angular velocity error under gyroscope saturations, especially for extreme values.
The spikes in estimated angular velocity are due to the free-fall assumptions which are no longer valid at the moment of collisions.

\subsection{Impact on SLAM}~\label{sec:result-slam}
\begin{figure}[htbp]
	\centering
	\includegraphics[width=\linewidth]{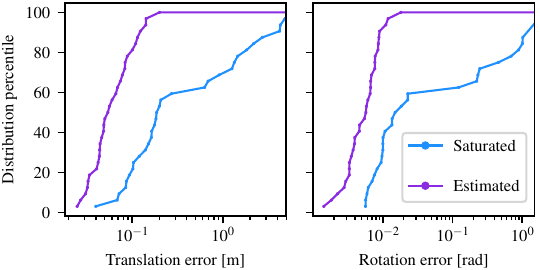}
	\caption{
            Localization error for every run in the dataset.
            The percentiles of translation error distribution is shown on the left plot and the percentiles of rotation error distribution is shown in the right plot.
            The blue and purple lines represent the percentiles of the localization error distribution when relying on saturated measurements and our speed estimation method, respectively.
            Errors on both subplots are shown with a log scale.}
	\label{fig:localization-error}
\end{figure}

Using no prior map of the environment, our \ac{SLAM} algorithm described in~\autoref{sec:slam-theory} was run on each of the \nbruns~runs in our dataset.
The localization errors with and without our angular velocity estimation method are illustrated in \autoref{fig:localization-error}. 
Here, the localization error corresponds to the error in the estimated transformation between the initial and final poses of the rig.
The ground-truth transformation for each run was found by registering the first and last scan in the ground-truth map, as the perception rig is static at these times.
Our angular velocity estimation approach improves the baseline~\ac{SLAM} algorithm localization error median by~\SI{71.5}{\%} for translation and~\SI{65.5}{\%} for rotation.

\begin{figure}[htbp]
	\centering
	\includegraphics[width=0.98\linewidth]{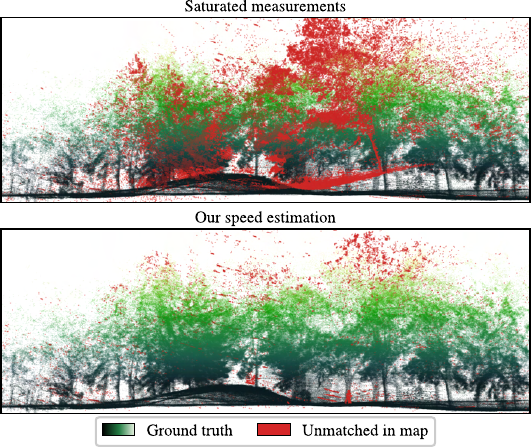}
        \caption{Side view of the ground-truth map built for the dataset.
        The color map is proportional to the point height.
        Mapping outliers from the fourteenth run are displayed in red.
        The top map shows the mapping outliers when relying on saturated measurements.
        The bottom map shows the mapping outliers when using our speed estimation method.
        Outlier points are defined as points that are farther than~\SI{0.25}{\meter} from the ground-truth map.
        }
	\label{fig:mapping-error}
\end{figure}

To investigate the impact of our method on mapping, we analyze a map built with our \ac{SLAM} system for every run in our dataset.
We built on prior work from~\citet{Chung2023} for the~\ac{DARPA} Subterranean Challenge to evaluate mapping quality.
Our map overlap metric is the percentage of reconstructed map points that are within a threshold distance from a point belonging to the ground-truth map.
In the present case, we chose the threshold distance to be \SI{0.25}{\meter} as opposed to the~\SI{1}{\meter} from the work of~\citet{Chung2023} to reflect the much smaller scale of our experiments.
Indeed, the distance traveled in our runs is between~\SI{5}{\meter} and~\SI{10}{\meter}, compared to between~\SI{150}{\meter} and~\SI{250}{\meter} in the case of the \ac{DARPA} Challenge.
The mean overlap of the maps built without our angular velocity estimation method is \SI{77.2}{\%}, as opposed to \SI{92.1}{\%} with our method.
The result for a specific run is shown in~\autoref{fig:mapping-error}.
We selected this run since the increase in mapping performance was significant when our~\ac{SLAM} algorithm relied on our speed estimation method.
Additionally, we define mapping failures as cases where the percentage of outliers in the reconstructed map (\ie points farther than \SI{0.25}{\meter} from their closest neighbor in the ground-truth map) is above \SI{15}{\percent}.
With saturated gyroscope measurements, we observe a failure of the mapping for 12 out of the 32 runs, as opposed to no failure when relying on our speed estimation method.
This clearly shows that our angular velocity estimation method increases mapping robustness in the case of a robot tumbling down a hill.
\subsection{The~\ac{TIGS} Dataset}
\label{sec:dataset_result}
\begin{figure}[htbp]
	\centering
	\includegraphics[width=0.98\linewidth]{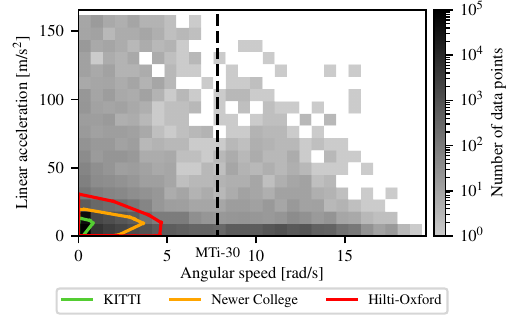}
        \caption{
        Density map of the~\ac{TIGS} dataset.
        The grayscale represents the number of data points acquired at the specific angular speeds and linear accelerations.
        The outlines represent the distributions in linear accelerations and angular velocities for similar datasets.
        The KITTI dataset is shown in green, the Newer College dataset is indicated in orange, and the Hilti-Oxford dataset is illustrated in red.
        The dashed line represents the manufacturer-specified saturation point of the MTi-30 gyroscope.
        }
	\label{fig:dataset_result}
\end{figure}

To show how our~\acf{TIGS} dataset covers a larger spectrum of aggressive motions than other mechanical lidar \ac{SLAM} datasets, we present the distributions for observed linear accelerations and angular velocities in~\autoref{fig:dataset_result}, in comparison to the KITTI~\citep{Geiger2012}, Newer College~\citep{Ramezani2020} and Hilti-Oxford~\citep{Zhang2023} datasets.
One can observe that our dataset covers a significantly larger spectrum of aggressive motions, characterized by high linear accelerations and angular velocities.
Indeed, the maximum recorded linear acceleration for the~\ac{TIGS} dataset is~\SI{\maxlinearaccelerationimprovement}{\meter/\square\second} over the highest linear acceleration observed in the compared datasets.
Since the saturation point of the VN-100 accelerometer was reached for some collisions, the increase in linear acceleration that was actually sustained is probably higher than this number.
Furthermore, the maximum angular speed for the~\ac{TIGS} dataset is~\SI{\maxangularspeedimprovement}{\radian/\second} over the highest angular speed observed in the compared datasets.
Our dataset is the only one with angular speeds over the specified saturation point of the Xsens gyroscope, thus allowing us to evaluate~\ac{SLAM} pipelines under saturated gyroscope measurements.

	\section{CONCLUSION}~\label{sec:conclu}
	In this paper, we introduced a novel method to estimate angular speed under saturated gyroscope measurements.
We validated our method through~\nbruns~runs mimicking a robot tumbling down a hill, with angular speeds reaching up to \SI{\maxangularspeed}{\radian/\second} and linear accelerations up to \SI{\maxlinearacceleration}{\meter/\square\second}.
Our system was able to perform \ac{SLAM} under these aggressive motions with no mapping failure, while it failed on \SI{37.5}{\percent} of the runs without our speed estimation method.
We release our dataset, called~\ac{TIGS}, to allow evaluation of~\ac{SLAM} frameworks under aggressive motions.
Future work involves quantifying the error of our method through a complete trajectory with a motion capture system and performing a thorough comparison with Point-LIO \cite{He2023}.

	
	%
	%
	\printbibliography

@article{Ebadi2023,
  title = {{Present and Future of SLAM in Extreme Environments: The DARPA SubT Challenge}},
  author = {Ebadi, Kamak and Bernreiter, Lukas and Biggie, Harel and Catt, Gavin and Chang, Yun and Chatterjee, Arghya and Denniston, Christopher E and Deschênes, Simon-Pierre and Harlow, Kyle and Khattak, Shehryar and Nogueira, Lucas and Palieri, Matteo and Petráček, Pavel and Petrlík, Matěj and Reinke, Andrzej and Krátký, Vít and Zhao, Shibo and Agha-Mohammadi, Ali-Akbar and Alexis, Kostas and Heckman, Christoffer and Khosoussi, Kasra and Kottege, Navinda and Morrell, Benjamin and Hutter, Marco and Pauling, Fred and Pomerleau, François and Saska, Martin and Scherer, Sebastian and Siegwart, Roland and Williams, Jason L and Carlone, Luca},
  doi = {10.1109/TRO.2023.3323938},
  journal = {IEEE Transactions on Robotics (T-RO)},
  month = {10},
  year = {2023},
}

@inproceedings{Dilaveroglu2020,
    author = {Dilaveroglu, Levent and Ozcan, Onur},
    doi = {10.1109/RoboSoft48309.2020.9115993},
    isbn = {9781728165707},
    booktitle = {3rd IEEE International Conference on Soft Robotics (RoboSoft)},
    title = {{MiniCoRe: A Miniature, Foldable, Collision Resilient Quadcopter}},
    year = {2020}
}

@article{Nagatani2013,
author = {Nagatani, Keiji and Kiribayashi, Seiga and Okada, Yoshito and Otake, Kazuki and Yoshida, Kazuya and Tadokoro, Satoshi and Nishimura, Takeshi and Yoshida, Tomoaki and Koyanagi, Eiji and Fukushima, Mineo and Kawatsuma, Shinji},
title = {{Emergency response to the nuclear accident at the Fukushima Daiichi Nuclear Power Plants using mobile rescue robots}},
journal = {Journal of Field Robotics (JFR)},
volume = {30},
number = {1},
pages = {44-63},
doi = {https://doi.org/10.1002/rob.21439},
url = {https://onlinelibrary.wiley.com/doi/abs/10.1002/rob.21439},
eprint = {https://onlinelibrary.wiley.com/doi/pdf/10.1002/rob.21439},
year = {2013}
}

@article{Dang2014,
   author = {Quoc Dang and Young Suh},
   doi = {10.3390/s140508167},
   issn = {1424-8220},
   issue = {5},
   journal = {Sensors},
   month = {5},
   pages = {8167-8188},
   publisher = {MDPI AG},
   title = {{Sensor Saturation Compensated Smoothing Algorithm for Inertial Sensor Based Motion Tracking}},
   volume = {14},
   url = {http://www.mdpi.com/1424-8220/14/5/8167},
   year = {2014},
}

@article{Silic2020,
	author = {Matthew Silic and Kamran Mohseni},
	doi = {10.1109/JSEN.2019.2945201},
	issn = {15581748},
	issue = {2},
	journal = {IEEE Sensors Journal},
	month = {1},
	pages = {1067-1076},
	title = {{Correcting Current-Induced Magnetometer Errors on UAVs: An Online Model-Based Approach}},
	volume = {20},
	year = {2020},
}

@article{Pachter2013,
   author = {Meir Pachter and Troy C. Welker and Richard E. Huffman},
   doi = {10.1002/navi.32},
   issn = {00281522},
   number = {2},
   journal = {{NAVIGATION}},
   pages = {85-96},
   publisher = {Wiley-Blackwell},
   title = {{Gyro-free INS Theory}},
   volume = {60},
   year = {2013},
}

@article{Lee2019,
   author = {Junhak Lee and Heyone Kim and Sang Heon Oh and Jae Chul Do and Chang Woo Nam and Dong Hwan Hwang and Sang Jeong Lee},
   doi = {10.1007/s00542-018-4281-8},
   issn = {09467076},
   issue = {7},
   journal = {Microsystem Technologies},
   month = {7},
   pages = {2855-2867},
   publisher = {Springer Verlag},
   title = {{Angular velocity estimation of rotating plate using extended Kalman filter with accelerometer bias model}},
   volume = {25},
   year = {2019},
}

@inproceedings{Tan2020,
   author = {Chee How Tan and Danial Sufiyan bin Shaiful and Emmanuel Tang and Jien-Yi Khaw and Gim Song Soh and Shaohui Foong},
   doi = {10.1109/ICRA40945.2020.9197486},
   isbn = {978-1-7281-7395-5},
   booktitle = {2020 IEEE International Conference on Robotics and Automation (ICRA)},
   month = {5},
   pages = {8532-8537},
   title = {{Flydar: Magnetometer-based High Angular Rate Estimation during Gyro Saturation for SLAM}},
   url = {https://ieeexplore.ieee.org/document/9197486/},
}

@article{Xu2022,
   author = {Wei Xu and Yixi Cai and Dongjiao He and Jiarong Lin and Fu Zhang},
   doi = {10.1109/TRO.2022.3141876},
   issn = {19410468},
   issue = {4},
   journal = {IEEE Transactions on Robotics (T-RO)},
   month = {8},
   pages = {2053-2073},
   publisher = {Institute of Electrical and Electronics Engineers Inc.},
   title = {{FAST-LIO2: Fast Direct LiDAR-Inertial Odometry}},
   volume = {38},
   year = {2022},
}

@inproceedings{Chen2023,
   author = {Kenny Chen and Ryan Nemiroff and Brett T. Lopez},
   doi = {10.1109/ICRA48891.2023.10160508},
   isbn = {979-8-3503-2365-8},
   booktitle = {2023 IEEE International Conference on Robotics and Automation (ICRA)},
   month = {5},
   pages = {3983-3989},
   title = {{Direct LiDAR-Inertial Odometry: Lightweight LIO with Continuous-Time Motion Correction}},
   url = {https://ieeexplore.ieee.org/document/10160508/},
}

@article{Palieri2021,
  author={Palieri, Matteo and Morrell, Benjamin and Thakur, Abhishek and Ebadi, Kamak and Nash, Jeremy and Chatterjee, Arghya and Kanellakis, Christoforos and Carlone, Luca and Guaragnella, Cataldo and Agha-mohammadi, Ali-akbar},
  journal={IEEE Robotics and Automation Letters (RA-L)}, 
  title={{LOCUS: A Multi-Sensor Lidar-Centric Solution for High-Precision Odometry and 3D Mapping in Real-Time}}, 
  year={2021},
  volume={6},
  number={2},
  pages={421-428},
  doi={10.1109/LRA.2020.3044864}
}

@article{Reinke2022,
   author = {Andrzej Reinke and Matteo Palieri and Benjamin Morrell and Yun Chang and Kamak Ebadi and Luca Carlone and Ali Akbar Agha-Mohammadi},
   doi = {10.1109/LRA.2022.3181357},
   issn = {23773766},
   issue = {4},
   journal = {IEEE Robotics and Automation Letters (RA-L)},
   month = {10},
   pages = {9043-9050},
   publisher = {Institute of Electrical and Electronics Engineers Inc.},
   title = {{LOCUS 2.0: Robust and Computationally Efficient Lidar Odometry for Real-Time 3D Mapping}},
   volume = {7},
   year = {2022},
}

@inproceedings{Shan2020,
   author = {Tixiao Shan and Brendan Englot and Drew Meyers and Wei Wang and Carlo Ratti and Daniela Rus},
   doi = {10.1109/IROS45743.2020.9341176},
   isbn = {9781728162126},
   issn = {21530866},
   booktitle = {2020 IEEE/RSJ International Conference on Intelligent Robots and Systems (IROS)},
   month = {10},
   pages = {5135-5142},
   title = {{LIO-SAM: Tightly-coupled lidar inertial odometry via smoothing and mapping}},
   }

@inproceedings{Madgwick2011,
    author={Madgwick, Sebastian O. H. and Harrison, Andrew J. L. and Vaidyanathan, Ravi},
    booktitle={2011 IEEE International Conference on Rehabilitation Robotics},
    title={{Estimation of IMU and MARG orientation using a gradient descent algorithm}},
    volume={},
    number={},
    pages={1-7},
    doi={10.1109/ICORR.2011.5975346}
}

@inproceedings{Deschenes2021,
	author={Deschênes, Simon-Pierre and Baril, Dominic and Kubelka, Vladimír and Giguère, Philippe and Pomerleau, François},
	booktitle={2021 18th Conference on Robots and Vision (CRV)},
	title={{Lidar Scan Registration Robust to Extreme Motions}},
	year={2021},
    publisher={IEEE},
	volume={},
	number={},
	pages={17-24},
	doi={10.1109/CRV52889.2021.00014}
}

@inproceedings{Segal2009,
   author = {A. Segal and D. Haehnel and S. Thrun},
   doi = {10.15607/RSS.2009.V.021},
   isbn = {9780262514637},
   booktitle = {Robotics: Science and Systems (RSS) V},
   publisher = {Robotics: Science and Systems Foundation},
   title = {{Generalized-ICP}},
   url = {http://www.roboticsproceedings.org/rss05/p21.pdf},
   year = {2009},
}

@inproceedings{Anderson2015,
	author = {Sean Anderson and Timothy D. Barfoot},
	doi = {10.1109/IROS.2015.7353368},
	isbn = {978-1-4799-9994-1},
	booktitle = {2015 IEEE/RSJ International Conference on Intelligent Robots and Systems (IROS)},
	month = {9},
	pages = {157-164},
	title = {{Full STEAM ahead: Exactly sparse gaussian process regression for batch continuous-time trajectory estimation on SE(3)}},
	url = {http://ieeexplore.ieee.org/document/7353368/},
}

@article{Baril2022,
author = {Baril, Dominic and Desch{\^{e}}nes, Simon-Pierre and Gamache, Olivier and Vaidis, Maxime and LaRocque, Damien and Laconte, Johann and Kubelka, Vladim{\'{i}}r and Gigu{\`{e}}re, Philippe and Pomerleau, Fran{\c{c}}ois},
doi = {10.55417/fr.2022050},
eprint = {2111.13981},
issn = {27713989},
journal = {Field Robotics},
month = {3},
number = {1},
pages = {1628--1660},
title = {{Kilometer-scale autonomous navigation in subarctic forests: challenges and lessons learned}},
url = {http://arxiv.org/abs/2111.13981 https://fieldrobotics.net/Field_Robotics/Volume_2_files/Vol2_50.pdf},
volume = {2},
year = {2022}
}

@article{Chung2023,
    author = {Chung, Timothy H. and Orekhov, Viktor and Maio, Angela},
    title = {{Into the Robotic Depths: Analysis and Insights from the DARPA Subterranean Challenge}},
    journal = {Annual Review of Control, Robotics, and Autonomous Systems},
    volume = {6},
    number = {1},
    pages = {477-502},
    year = {2023},
    doi = {10.1146/annurev-control-062722-100728},
}

@inproceedings{Ramezani2020,
	author = {Milad Ramezani and Yiduo Wang and Marco Camurri and David Wisth and Matias Mattamala and Maurice Fallon},
	doi = {10.1109/IROS45743.2020.9340849},
	isbn = {978-1-7281-6212-6},
	issn = {21530866},
	booktitle = {2020 IEEE/RSJ International Conference on Intelligent Robots and Systems (IROS)},
	month = {10},
	pages = {4353-4360},
	title = {{The Newer College Dataset: Handheld LiDAR, Inertial and Vision with Ground Truth}},
	url = {https://ieeexplore.ieee.org/document/9340849/},
}

@inproceedings{Geiger2012,
   author = {A. Geiger and P. Lenz and R. Urtasun},
   doi = {10.1109/CVPR.2012.6248074},
   isbn = {978-1-4673-1228-8},
   booktitle = {2012 IEEE Conference on Computer Vision and Pattern Recognition},
   pages = {3354-3361},
   title = {{Are we ready for autonomous driving? The KITTI vision benchmark suite}},
   url = {http://ieeexplore.ieee.org/document/6248074/},
}

@article{Zhang2023,
   author = {Lintong Zhang and Michael Helmberger and Lanke Frank Tarimo Fu and David Wisth and Marco Camurri and Davide Scaramuzza and Maurice Fallon},
   doi = {10.1109/LRA.2022.3226077},
   issn = {2377-3766},
   issue = {1},
   journal = {IEEE Robotics and Automation Letters (RA-L)},
   month = {1},
   pages = {408-415},
   title = {{Hilti-Oxford Dataset: A Millimeter-Accurate Benchmark for Simultaneous Localization and Mapping}},
   volume = {8},
   url = {https://ieeexplore.ieee.org/document/9968057/},
   year = {2023},
}

@article{Williams2018,
	author = {Williams, Grady and Drews, Paul and Goldfain, Brian and Rehg, James M. and Theodorou, Evangelos A.},
	doi = {10.1109/TRO.2018.2865891},
	issn = {1552-3098},
	journal = {IEEE Transactions on Robotics (T-RO)},
	month = {12},
	number = {6},
	pages = {1603--1622},
	title = {{Information-Theoretic Model Predictive Control: Theory and Applications to Autonomous Driving}},
	url = {https://ieeexplore.ieee.org/document/8558663/},
	volume = {34},
	year = {2018}
}

@ARTICLE{Tang2019,
  author={Tang, Tim Yuqing and Yoon, David Juny and Barfoot, Timothy D.},
  journal={IEEE Robotics and Automation Letters}, 
  title={{A White-Noise-on-Jerk Motion Prior for Continuous-Time Trajectory Estimation on SE(3)}}, 
  year={2019},
  volume={4},
  number={2},
  pages={594-601},
  doi={10.1109/LRA.2019.2891492}
}

@article{He2023,
   author = {Dongjiao He and Wei Xu and Nan Chen and Fanze Kong and Chongjian Yuan and Fu Zhang},
   doi = {10.1002/aisy.202200459},
   issn = {2640-4567},
   issue = {7},
   journal = {Advanced Intelligent Systems},
   month = {7},
   publisher = {John Wiley and Sons Inc},
   title = {{Point‐LIO: Robust High‐Bandwidth Light Detection and Ranging Inertial Odometry}},
   volume = {5},
   url = {https://onlinelibrary.wiley.com/doi/10.1002/aisy.202200459},
   year = {2023},
}
	
\end{document}